\documentclass[letterpaper]{article} 
\usepackage[submission]{aaai23}  
\usepackage{times}  
\usepackage{helvet}  
\usepackage{courier}  
\usepackage[hyphens]{url}  
\usepackage{graphicx} 
\usepackage{natbib}  
\usepackage{caption} 
\usepackage{algorithm}
\usepackage{algorithmic}

\usepackage{newfloat}
\usepackage{listings}
\DeclareCaptionStyle{ruled}{labelfont=normalfont,labelsep=colon,strut=off} 
\floatstyle{ruled}
\newfloat{listing}{tb}{lst}{}
\floatname{listing}{Listing}
\usepackage{kotex}
\usepackage{makecell}
\usepackage{amsfonts}       
\usepackage{amsmath}
\usepackage{multirow}
\usepackage{subcaption}
\usepackage{tabularx}
\usepackage{enumitem}
\usepackage{booktabs}
\usepackage{soul}
\usepackage{xcolor}

\title{Meta-node: A Concise Approach to \\Effectively Learn Complex Relationships in Heterogeneous Graphs}

\author{
    Jiwoong Park \textsuperscript{\rm 1}
    Jisu Jeong \textsuperscript{\rm 2}
    Kyungmin Kim \textsuperscript{\rm 2}
    Jin Young Choi \textsuperscript{\rm 1}
}
\affiliations{
    \textsuperscript{\rm 1}Seoul National University,
    \textsuperscript{\rm 2}NAVER

}

\begin{document}

\maketitle

\begin{abstract}
Existing message passing neural networks for heterogeneous graphs rely on the concepts of meta-paths or meta-graphs due to the intrinsic nature of heterogeneous graphs.
However, the meta-paths and meta-graphs need to be pre-configured before learning and are highly dependent on expert knowledge to construct them. 
To tackle this challenge, we propose a novel concept of \textit{meta-node} for message passing that can learn enriched relational knowledge from complex heterogeneous graphs without any meta-paths and meta-graphs by explicitly modeling the relations among the same type of nodes.
Unlike meta-paths and meta-graphs, meta-nodes do not require any pre-processing steps that require expert knowledge.
Going one step further, we propose a meta-node message passing scheme and apply our method to a contrastive learning model.
In the experiments on node clustering and classification tasks, the proposed meta-node message passing method outperforms state-of-the-arts that depend on meta-paths.
Our results demonstrate that effective heterogeneous graph learning is possible without the need for meta-paths that are frequently used in this field.
\end{abstract}
\section{Introduction}
Graph Neural Networks (GNNs) \cite{gori2005new, gilmer2017neural, kipf2017semi, hamilton2017inductive, velickovic2018graph, xu2018representation} have become the de facto standard for representation learning on graph-structured data.
\begin{figure}[t]
  \centering
  \includegraphics[clip,trim={0 5cm 0 0}, width=\linewidth]{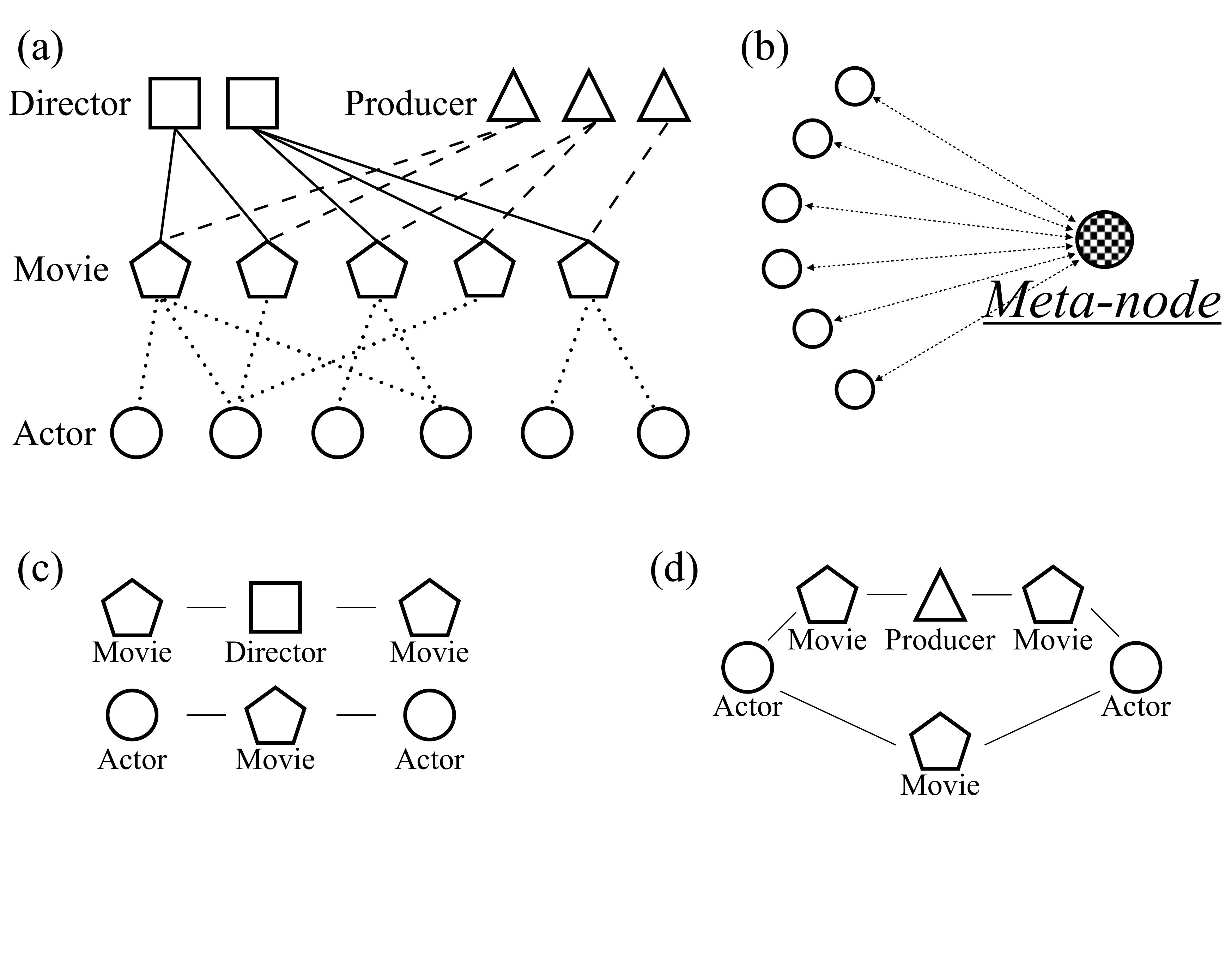}
  \caption{
  (a) An example of heterogeneous movie networks. There exists four types of nodes: \textit{movie, actor, director}, and \textit{producer}. 
  (b) The proposed meta-node: each meta-node aggregates messages of all nodes of each type and returns the aggregated message to each node when passing messages to the next layer.
  (c) Example of two meta-paths which are compositions of different types of nodes. 
  (d) A meta-graph which is a composition of multiple meta-paths.
  }
  \label{movie_network}
  \vskip -0.15in
\end{figure}
Among the differing architectures for GNNs, Message Passing Neural Networks (MPNNs) \cite{gilmer2017neural, morris2019weisfeiler} in which nodes exchange messages (i.e., representations) along edges, are considered well-known and effective mechanisms.
Since GNNs were first proposed \cite{gori2005new}, the majority of efforts in this field have been aimed at learning representations for homogeneous graphs with a single type of nodes and a single type of edges (i.e., relationships).

However, graph-structured datasets in real-world applications are not limited to a single type of nodes and edges.
For instance, in the movie network of Figure \ref{movie_network} (a), there exist multiple types of nodes (movies, actors, directors, and producers) and multiple types of edges (acting, filming, and producing).
This kind of graph that has multiple types of nodes and edges is called \textit{heterogeneous graph} \cite{wang2020survey, yang2020heterogeneous}.
To capture complex relations in heterogeneous graphs, the representation learning model must consider the distinct nature of multiple types of nodes and edges.
Thus simply plugging heterogeneous graphs into conventional MPNNs 
is inadequate because the MPNNs cannot distinguish multiple node and edge types.
To deal with this problem, recently, Heterogeneous Graph Neural Networks (HGNNs) \cite{schlichtkrull2018modeling, zhang2019heterogeneous, wang2019heterogeneous, kim2019tripartite, yun2019graph, wang2021heco, ren2020hdgi, hu2020heterogeneous} have been proposed to extract useful knowledge from heterogeneous graphs by leveraging the power of GNNs.
\begin{table*}[ht]
  \caption{Predefined meta-paths of real-world datasets.
  In this table, it can be noticed that most of $\mathcal{R}$ are inter-type relations and $\mathcal{P}$ target on intra-type relations by setting the same type of nodes at both ends of $\mathcal{P}$.}
  \label{tab:meta_path}
  \resizebox{0.99\textwidth}{!}{\begin{tabular}{llll}
    \toprule
    Dataset & $\mathcal{A}$ & $\mathcal{R}$ & $\mathcal{P}$\\
    \midrule
    DBLP & A, P, T, C & A-P, P-T, P-C & APA, APCPA, APTPA\\
    IMDB & M, D, A & M-D, M-A & MDM, MAM\\
    ACM & P, A, S & P-A, P-S & PAP, PSP \\
    AMiner & P, A, R & P-A, P-R & PAP, PRP\\
    Freebase & M, D, A, P & M-D, M-A, M-P & MAM, MDM, MPM \\
    Last.FM & U, A, T & U-U, U-A, A-T & UU, UAU, UATAU, AUA, AUUA, ATA\\
    Yelp & U, B, Co, Ci, Ca & U-U, U-B, U-Co, B-Ci, B-Ca & UBU, UCoU, UBCiBU, UBCaBU, BUB, BCiB, BCaB, BUCoUB \\
    Douban & U, M, G, L, D, A, T & U-U, U-G, U-M, U-L, M-D, M-T, M-A & MUM, MTM, MDM, MAM, UMU, UMAMU, UMDMU, UMTMU \\
  \bottomrule
\end{tabular}}
\end{table*}

To learn complex relational knowledge from heterogeneous graphs, most HGNNs rely on the node compositions constructed before training.
This dependency on pre-processing steps of HGNNs is from the unique characteristics of heterogeneous graphs.
As an example in Figure \ref{movie_network}, most of the heterogeneous graphs are $k$-partite graphs whose nodes can be divided into $k$ independent sets.
Due to the nature of $k$-partite graphs, all that is given are sparse inter-type relations (i.e., edges between different types of nodes).
However, using only these inter-type relations is not enough to extract useful knowledge from the intricate relations in the data.
To resolve this problem, most HGNNs rely on additional predefined relational information, and the most commonly used methods are \textit{meta-path} \cite{sun2011pathsim} and \textit{meta-graph} \cite{fang2016semantic, huang2016meta}, each of which are a composition of different types of nodes and multiple meta-paths as shown in Figure \ref{movie_network} (c) and (d).
As we will show later, nearly all meta-paths implicitly derive intra-type relations (i.e., relations between the same type of nodes) by manipulating given inter-type relations.

However, there exist three major problems with using predefined methods such as meta-paths for heterogeneous graph learning.
Firstly, there exist certain limitations on inducing intra-type relations from predefined inter-type relations.
When the given inter-type relations are sparse or noisy, induced intra-type relations can also be affected.
Secondly, the appropriate composition of nodes and edges (designing meta-paths and meta-graphs) for representation learning requires significant domain-specific knowledge.
Thus, it is extremely hard to know which combinations of nodes and edges are suitable for learning useful representations, especially in unsupervised environments.
Lastly, although there exist attempts to learn appropriate meta-paths beyond given ones \cite{yun2019graph}, several multiplications of the adjacency matrix are required. 
Due to the high computational cost of multiple matrix multiplications, their method is limited to very small datasets \cite{lv2021are}.

To circumvent the above limitations of current methods, we propose a novel concept of \textbf{\textit{meta-node}} to construct simple and powerful MPNNs for learning heterogeneous graphs.
Meta-nodes are virtual nodes in which one meta-node is added to the graph for each type of node in the heterogeneous graph.
Each meta-node is connected to all nodes of each type as illustrated in Figure \ref{movie_network} (b).
By introducing meta-nodes, message passing is no longer limited to sparse inter-type relations, and every node can directly perform message passing with other nodes of the same type via meta-nodes.
To do so, we can enrich the information on the relationship by adding explicit intra-type relations to the given inter-type relations.
After introducing the concept of meta-nodes, we propose a message passing scheme via meta-node to learn both intra- and inter-type relations effectively.

Unsupervised representation learning on heterogeneous graphs has become one of the major challenges in  graph-structured data learning, as it can pave the way to make use of large amounts of unlabeled multi-modal data.
Thus, we validate the proposed message passing scheme by applying it to unsupervised representation learning for graph-structured data.
To do so, we apply our meta-node message passing layer to the encoder of Deep Graph Infomax \cite{velickovic2018deep} which is one of the most well-known graph contrastive models.
Through downstream tasks on four real-world heterogeneous graph datasets, we validate the proposed message passing scheme.
We confirm that our meta-node message passing layer learns rich relational information and shows competitive performance compared to existing state-of-the-art HGNNs even without any meta-paths.

\section{Related Work}
\subsubsection{Meta-path.} 
A meta-path \cite{sun2011pathsim} $\mathcal{P}$ is defined as a path that has a form of $A_1 \xrightarrow{\mbox{\tiny{$R_1$}}} A_2 \xrightarrow{\mbox{\tiny{$R_2$}}} \cdots \xrightarrow{\mbox{\tiny{$R_l$}}} A_{l+1}$ (abbreviated as $A_1 A_2 \cdots A_{l+1}$) which describes relations between $A_1$ and $A_{l+1} \in \mathcal{A}$ with a composition of relations $R_1, R_2, \ldots ,R_l \in \mathcal{R}$, where $\mathcal{A}$ and $\mathcal{R}$ denote sets of node types and edge types of heterogeneous graphs, respectively. 
Each meta-path can describe a semantic relation between nodes at both ends of the meta-path.
For instance, in Figure \ref{movie_network} (c), the meta-path of movie-director-movie can describe the relationship between two movies by which the director filmed them.
Nearly all meta-paths of the real-world datasets \cite{wang2019heterogeneous,fu2020magnn, wang2020survey, wang2021heco} are implicitly composed for intra-type relations by setting the same type of nodes at both ends of $\mathcal{P}$ using given inter-type relations $\mathcal{R}$ as shown in Table \ref{tab:meta_path}.
\begin{figure*}[t]
  \centering
  \includegraphics[clip,trim={0 2cm 0 0}, width=\linewidth]{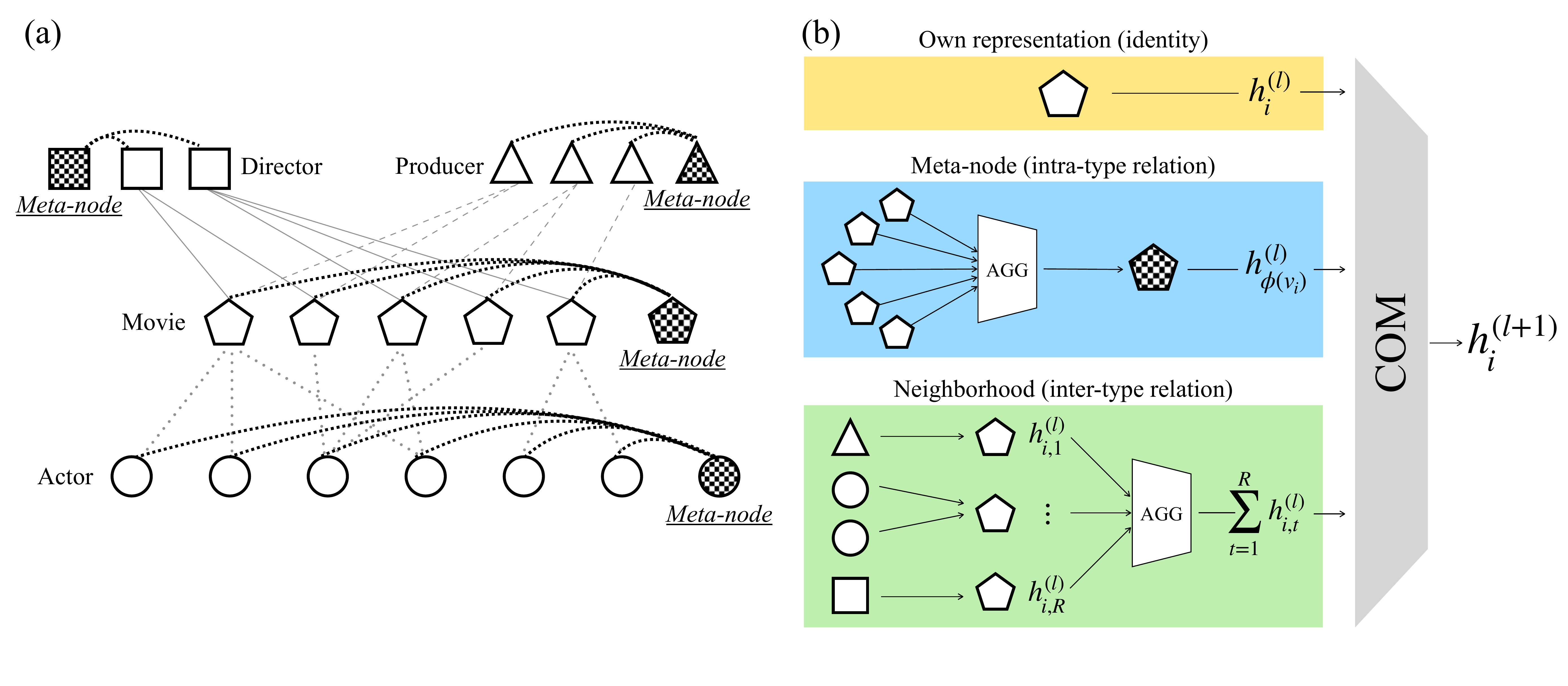}
  \caption{(a) The concept of meta-nodes is schematically visualized.
  Each meta-node (checkerboard pattern) is connected to all nodes of one node type using an extended set of edges (black dashed lines). 
  (b) The proposed meta-node message passing layer (MN-MPL) takes three components as in Eq. (\ref{message}): its own representation from the previous layer (yellow), the intra-type relations derived from the meta-node representation aggregated from the same type nodes (blue), and the inter-type relations derived from the neighborhood messages aggregated from direct heterogeneous neighbors (green).
  }
  \label{fig:meta_node}
\end{figure*}

\subsubsection{Representation Learning for Heterogeneous Graphs.} 
For several past years, there have been many efforts to learn representations of heterogeneous graphs based on random-walk-based methods \cite{dong2017metapath2vec, fu2017hin2vec, jeong2020div2vec, he2019hetespaceywalk} or GNNs methods \cite{schlichtkrull2018modeling, shi2018heterogeneous, zhang2019heterogeneous, yun2019graph, wang2019heterogeneous, zhao2020network, fu2020magnn, hu2020heterogeneous, zhao2021heterogeneous}.
Nowadays, HGNNs leveraging the power of GNNs show a remarkable ability to learn intricate relations between multiple types of nodes and edges both in semi-supervised and unsupervised conditions.
For instance, in semi-supervised learning, HAN \cite{zhang2019heterogeneous} proposed attention-based MPNNs using meta-paths to take each semantic meaning of meta-paths into account.
Further, MAGNN \cite{fu2020magnn} considered all the nodes constituting each meta-path to respect intermediate semantic information.
GTN \cite{yun2019graph} is an advanced method to learn new relations (meta-paths) in heterogeneous graphs by several multiplications of the adjacency matrix.
In unsupervised learning, HetGNN \cite{zhang2019heterogeneous} learns a fixed size of correlated neighbors using random walks with restart.
NSHE \cite{zhao2020network} samples subgraphs and learns via multi-task learning to preserve relations in heterogeneous graphs.

\subsubsection{Contrastive Methods.}
Starting from a pioneering work of contrastive learning for homogeneous graphs \cite{velickovic2018deep}, 
there are some efforts to learn heterogeneous graphs through contrastive methods.
HDMI \cite{jing2021hdmi} utilizes higher-order mutual information via considering relations of raw features of nodes, learned representations, and global summary vectors.
HDGI \cite{ren2020hdgi} is a contrastive model whose encoder is meta-path-based neighborhood aggregation and contrasts between an original graph and a corrupted graph.
HeCo \cite{wang2021heco} is another contrastive model that contrasts local representation results of direct heterogeneous neighborhood aggregation and the representation to those of meta-path-based neighborhood aggregation.

\subsubsection{Limitations of Existing Works.} Whether the target condition of models is semi-supervised or unsupervised, it can be noticed that many heterogeneous graph learning models rely on predefined meta-paths or meta-graphs.
However, depending on predefined meta-paths is undesirable since the results of meta-paths based message passing cannot directly leverage the relationships between nodes of the same type.
Also, it is difficult to know whether meta-paths given in advance are conducive to effective learning.
Thus, the question of how to effectively learn node representations of heterogeneous graphs without any predefined relational knowledge is one of the major challenges to be solved.

\section{Proposed Method}
In this section, we present the meta-node concept, a meta-node message passing layer (MN-MPL), and a contrastive model for heterogeneous graphs.
The encoder of the contrastive model adopts a MN-MPL.
The graphical descriptions of the meta-node, MN-MPL, and the contrastive model are illustrated in Figures \ref{fig:meta_node} and \ref{fig:contrastive_model}.

\subsection{Definitions}
\subsubsection{Definition 1. Heterogeneous graph.} 
A heterogeneous graph $G = (\mathcal{V}, \mathcal{E}, \mathcal{A}, \mathcal{R}, \phi, \psi)$ is a network with multiple types of nodes and edges, where $\mathcal{V}$ and $\mathcal{E}$ denote the node set and edge set, respectively.
Each node $v \in \mathcal{V}$ and edge $e \in \mathcal{E}$ are associated with a node type mapping function $\phi(v): \mathcal{V} \rightarrow \mathcal{A}$ and an edge type mapping function $\psi(e): \mathcal{E} \rightarrow \mathcal{R}$, where $\mathcal{A}$ and $\mathcal{R}$ denote sets of node types and edge types with $|\mathcal{A}| + |\mathcal{R}| > 2$, respectively. 
$\mathcal{V}_j$ denotes the node set having $j$-th node type, that is, $\bigcup_{j=1}^{|\mathcal{A}|} \mathcal{V}_j = \mathcal{V}$.

\subsubsection{Definition 2. Meta-node.} 
Meta-nodes $\{\mathbf{v}_j \}_{j=1}^{|\mathcal{A}|}$ are additional nodes whereby one node is introduced for each node type in $G$.
Each meta-node $\mathbf{v}_j$ is connected to all nodes in $\mathcal{V}_j$ via added edges that connect each meta-node and the nodes in each node type.
The added edges enable message passing between each meta-node $\mathbf{v}_j$ and all nodes in $\mathcal{V}_j$.

We illustrate an example of a heterogeneous graph introducing meta-nodes in Figure \ref{fig:meta_node} (a).
Introducing the concept of meta-nodes enables explicit modeling of intra-type relations that is otherwise hard to infer due to the $k$-partite structural characteristics of heterogeneous graphs.
Compared to conventional methods of using meta-paths or meta-graphs that infer intra-type relations from given inter-type relations indirectly, \textit{meta-nodes can directly establish intra-type relations.}
Also, unlike meta-paths or meta-graphs that are predefined before learning, \textit{meta-nodes do not require any prior domain knowledge or predefined steps.}

\subsection{MN-MPL: Meta-node Message Passing Layer}
We now propose a novel message passing layer using meta-nodes.
The proposed layer takes three components as input: the representation of the previous layer, the meta-node representation, and aggregated messages from direct heterogeneous neighbors as shown in Figure \ref{fig:meta_node} (b).
For a node $v_i \in \mathcal{V}$ who has $R$ different types of immediate neighbors, the meta-node message passing layer (MN-MPL) can be expressed as
\begin{equation}
  h_i^{(l+1)} = \text{COM}\Big(h_i^{(l)},  h^{(l)}_{\phi(v_i)}, 
  \sum_{t=1}^R  h_{i,t}^{(l)} \Big),
  \label{message}
\end{equation}
where $h_i^{(l)}, h^{(l)}_{\phi(v_i)}$, and $h_{i,t}^{(l)}$ denote the representation of the $i$-th node at the $l$-th layer, the meta-node representation that represents the type of the $i$-th node $\phi(v_i)$, and the aggregated representation from $t$ type neighbors of $v_i$, respectively.
For the combination function COM, we apply concatenation $\|_{j=1}^k x_j$ or summation $\sum_{j=1}^k x_j$. 
$\text{COM}$ includes a nonlinear MLP after concatenation or summation.
The meta-node representation $h^{(l)}_{\phi(v_i)}$ is defined by sum pooling: $\sum_{j \in \mathcal{V}_{\phi(v_i)}}h_j^{(l)}$, mean pooling: $\frac{1}{|\mathcal{V}_{\phi(v_i)}|}\sum _{j \in \mathcal{V}_{\phi(v_i)}}h_j^{(l)}$, or max pooling: $\max \{h_j^{(l)}\}_{j \in \mathcal{V}_{\phi(v_i)}}$, where $\max$ denotes the element-wise max function.
To aggregate messages from $R$ types of direct heterogeneous neighbors of $v_i$, we aggregate messages of each different type of direct neighbors separately first $\{ h_{i,1}^{(l)}, \ldots, h_{i,R}^{(l)} \}$, then sum them to make a single vector representation: $\sum_{t=1}^R  h_{i,t}^{(l)}$.

\subsection{Advantages of MN-MPL}
The proposed MN-MPL has three major advantages over conventional message passing on heterogeneous graphs.
Firstly, each node can exchange messages with nodes of the same type by taking the aggregated messages through meta-nodes as input during the proposed message passing process.
Thus, the proposed message passing scheme makes full use of both inter-type relations via direct heterogeneous neighbors and intra-type relations via meta-node representations.
Compared to conventional message passing schemes that infer intra-type relations indirectly using predefined meta-paths or meta-graphs, the proposed layer does not require any indirect infer or pre-processing steps before learning.
Secondly, through the meta-nodes, each node can easily exchange messages with distant nodes without having to pass through as many layers as the length of the meta-path.
With our newly introduced meta-nodes, since each meta-node connects all nodes of each type, nodes within each type can consider the others as one-hop neighbors.
Thus, classically distant but informative nodes can be learned with only a small number of MN-MPLs.
Lastly, the cost of message passing between intra-type nodes is extremely low.
As explained in the previous subsection, the computation of meta-node representations is a simple sum, mean or max pooling operation on the node representations, and infers negligible computational cost.
Thus, the computational cost of establishing unseen relations of heterogeneous graphs using meta-nodes is extremely low compared to existing methods such as \cite{yun2019graph} requiring several adjacency matrix multiplications that attempts to do an efficient variant very recently \cite{yun2021graph}.

\subsection{Contrastive Learning Framework}
\begin{figure}[t!]
  \centering
  \includegraphics[clip,trim={0 2cm 0 0}, width=0.9\linewidth]{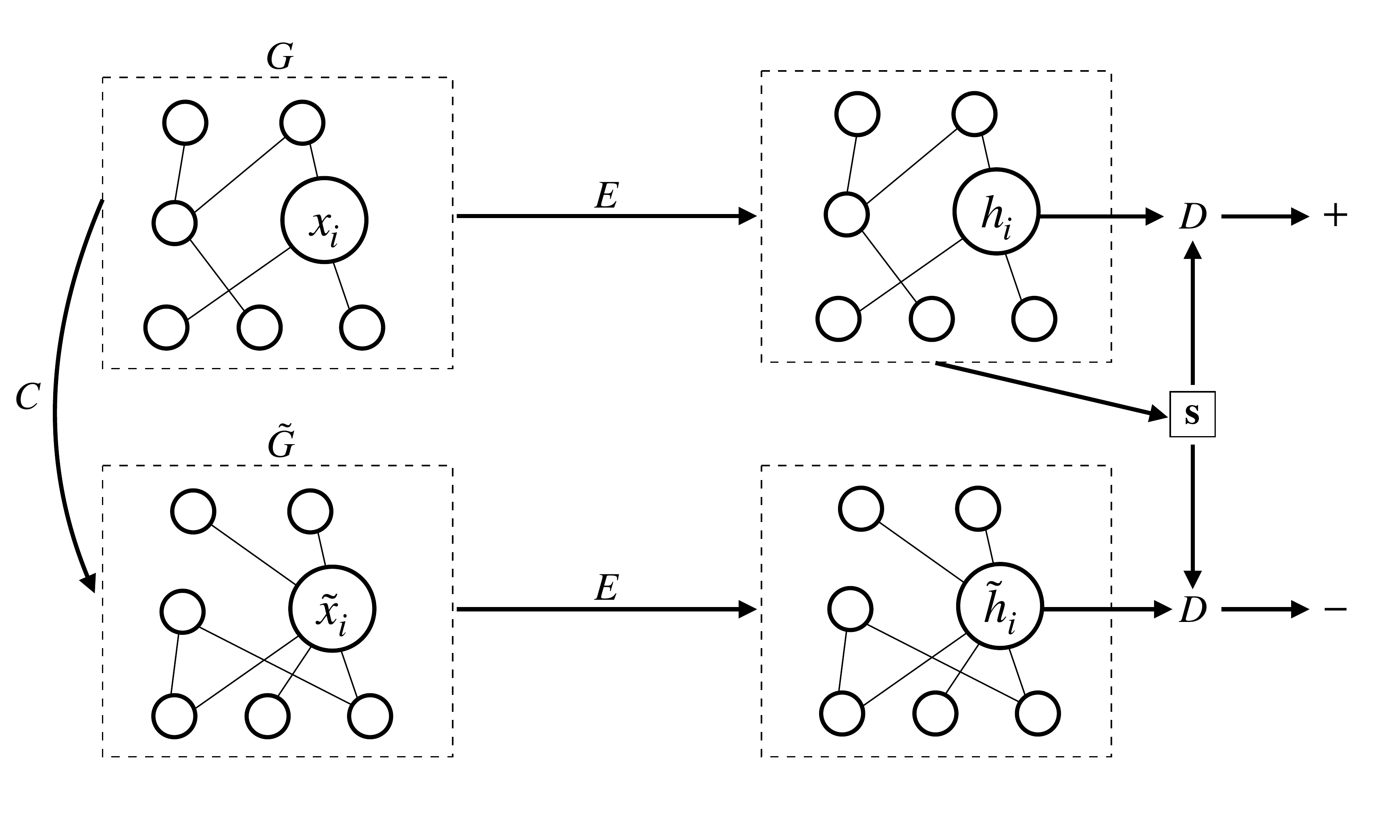}
  \caption{Overview of the contrastive model. $C, D, E$, and $\mathbf{s}$ denote a corruption function for generating negative samples, a discriminator function, an encoder network that is composed of our MN-MPL, and a global summary vector, respectively.
  We referred to the graphical description of \cite{velickovic2018deep}.}
  \label{fig:contrastive_model}
\end{figure}
We apply MN-MPLs to the contrastive framework of Deep Graph Infomax \cite{velickovic2018deep}.
At first, because each node type has attributes of different dimensions, we project each different attribute into a common latent space whose dimension is $d$ using one layer transformation network:
\begin{equation}
  h_i^{(0)} = \zeta (W_{\phi(v_i)}x_i + b_{\phi{(v_i)}}), 
  \label{transform}
\end{equation}
where $x_i \in \mathbb{R}^{d_{\phi(v_i)}}, W_{\phi(v_i)} \in \mathbb{R}^{d \times d_{\phi(v_i)}}, b_{\phi{(v_i)}} \in \mathbb{R}^d$, and $\zeta$ denote the node attribute of $v_i$, a transformation matrix, a bias vector for type $\phi(v_i)$, and the nonlinear activation, respectively.
For constrastive learning, we apply a corruption function $C$ to generate a negative graph $\tilde{G} = C(G)$.
We select the corruption function $C$ as a type-wise random permutation of the node feature matrix.
By applying $C$, each node is given the features of other nodes of the same type.

To learn the representation of each node, we apply MN-MPLs as the encoder network.
We share the same encoder network for both the original graph and the corrupted graph to learn the representation of each node.
$h_i$ and $\tilde{h}_i$ on node $v_i \in \mathcal{V}$ denote the outputs of the encoder network for the original graph and the corrupted graph, respectively.
We extract global summary vector $\mathbf{s}$ of the original graph by applying mean pooling $\mathbf{s} = \sigma( \frac{1}{|\mathcal{V}|}\sum_{i=1}^{|\mathcal{V}|} h_i)$, where $\sigma$ denotes the logistic sigmoid function.
Then, we utilize a contrastive objective with binary cross entropy loss function between positive samples $(h_i, \mathbf{s})$ and negative samples $(\tilde{h}_i, \mathbf{s})$:
\begin{equation}
  \small
  L = \frac{1}{2|\mathcal{V}|}\big(\sum_{i=1}^{|\mathcal{V}|}\mathbb{E}_G[\log \mathit{D}(h_i, \mathbf{s})] + \sum_{i=1}^{|\mathcal{V}|}\mathbb{E}_{\tilde{G}}[\log(1-\mathit{D}(\tilde{h}_i, \mathbf{s}))]\big),
  \label{loss}
\end{equation}
where $D(h_i, \mathbf{s}) = \sigma(h_i^T W \mathbf{s})$ denotes the discriminator function which is a bilinear network ($W$ is a learnable matrix).
Maximizing the objective function $L$ is equal to maximizing the mutual information between the representation from the original graph $h_i$ and the global summary vector $\mathbf{s}$ from the original graph. 
The graphical overview of the contrastive learning framework is illustrated in Figure \ref{fig:contrastive_model}.

\section{Experiments}
To verify the validity of the proposed meta-node message passing scheme, we applied the contrastive model with MN-MPL to node classification and node clustering tasks on the target node type of each dataset.
Further, we analyzed the quality of the learned representation of nodes and the effectiveness of the proposed method via additional analysis.

\subsection{Experimental Settings}
\subsubsection{Datasets.}
We validated the contrastive learning model based on our MN-MPL using four real-world heterogeneous graph datasets. 
The statistics of datasets are presented in Table \ref{tab:data}.
The details and download links of datasets are presented in the supplementary material.

\subsubsection{Compared Methods.} We compared our method with graph representation learning methods in three categories: i) \textit{unsupervised homogeneous models}: n2vec \cite{grover2016node2vec}, SAGE \cite{hamilton2017inductive}, GAE \cite{kipf2016variational}, DGI \cite{velickovic2018deep}, ii) \textit{unsupervised heterogeneous models}: mp2vec \cite{dong2017metapath2vec}, HERec \cite{shi2018heterogeneous}, HetGNN \cite{zhang2019heterogeneous}, DMGI \cite{park2020unsupervised}, HeCo \cite{wang2021heco}, and iii) \textit{semi-supervised heterogeneous model}: HAN \cite{wang2019heterogeneous}.
The brief explanation and the download link of each compared method are presented in the supplementary material.
\begin{table}[t]
  \caption{Statistics of datasets. The target node type of each dataset is shown in bold.}
  \centering
  \label{tab:data}
  \resizebox{0.4\textwidth}{!}{\begin{tabular}{lccc}
    \toprule
    Dataset & $\mathcal{A}$ & $\mathcal{R}$ & $\mathcal{P}$\\
    \midrule
     DBLP & \makecell*[c]{\textbf{Author (A): 4,057}\\Paper (P): 14,328\\Term (T): 7,723\\Conference (C): 20} & \makecell*[c]{A-P: 19,645\\P-T: 85,810\\P-C: 14,328} & \makecell*[c]{APA\\APTPA\\APCPA} \\
    \midrule
    ACM & \makecell*[c]{\textbf{Paper (P): 4,019}\\Author (A): 7,167\\Subject (S): 60} & \makecell*[c]{P-A: 13,407\\P-S: 4,019} & \makecell*[c]{PAP\\PSP} \\
    \midrule
    AMiner & \makecell*[c]{\textbf{Paper (P): 6,564}\\Author (A): 13,329\\Reference (R): 35,890} & \makecell*[c]{P-A: 18,007\\P-R: 58,831} & \makecell*[c]{PAP\\PRP} \\
    \midrule
    Freebase & \makecell*[c]{\textbf{Movie (M): 3,492}\\Director (D): 2,502\\Actor (A): 33,401\\Producer (P): 4,459} & \makecell*[c]{M-D: 3,762\\M-A: 65,341\\M-P: 6,414} & \makecell*[c]{MDM\\MAM\\MPM} \\
  \bottomrule
\end{tabular}}
\end{table}

\begin{table*}[ht]
  \caption{Summary of node classification results ($\%\pm\sigma$).}
  \label{tab:classification}
  \resizebox{\textwidth}{!}{\begin{tabular}{lccccccccccccc}
    \toprule
    Datasets & Metric & Split & n2vec & SAGE & GAE & mp2vec & HERec & HetGNN & HAN & DGI & DMGI & HeCo & MN (ours)\\
    \midrule
    \multirow{9}{*}{DBLP}&
    \multirow{3}{*}{Mac-F1}
    &20&48.75$\pm$1.0&71.97$\pm$8.4&90.90$\pm$0.1&88.98$\pm$0.2&89.57$\pm$0.4&89.51$\pm$1.1&89.31$\pm$0.9&87.93$\pm$2.4&89.94$\pm$0.4&91.28$\pm$0.2&\textbf{93.43$\pm$0.5}\\
    &&40&55.94$\pm$1.0&73.69$\pm$8.4&89.60$\pm$0.3&88.68$\pm$0.2&89.73$\pm$0.4&88.61$\pm$0.8&88.87$\pm$1.0&88.62$\pm$0.6&89.25$\pm$0.4&90.34$\pm$0.3&\textbf{92.47$\pm$0.4}\\
    &&60&58.15$\pm$0.7&73.86$\pm$8.1&90.08$\pm$0.2&90.25$\pm$0.1&90.18$\pm$0.3&89.56$\pm$0.5&89.20$\pm$0.8&89.19$\pm$0.9&89.46$\pm$0.6&90.64$\pm$0.3&\textbf{93.72$\pm$0.4}\\
    \cmidrule{2-14}
    &\multirow{3}{*}{Mic-F1}
    &20&48.92$\pm$1.0&71.44$\pm$8.7&91.55$\pm$0.1&89.67$\pm$0.1&90.24$\pm$0.4&90.11$\pm$1.0&90.16$\pm$0.9&88.72$\pm$2.6&90.78$\pm$0.3&91.97$\pm$0.2&\textbf{93.88$\pm$0.5}\\
    &&40&56.06$\pm$1.1&73.61$\pm$8.6&90.00$\pm$0.3&89.14$\pm$0.2&90.15$\pm$0.4&89.03$\pm$0.7&89.47$\pm$0.9&89.22$\pm$0.5&89.92$\pm$0.4&90.76$\pm$0.3&\textbf{92.79$\pm$0.4}\\
    &&60&58.58$\pm$0.8&74.05$\pm$8.3&90.95$\pm$0.2&91.17$\pm$0.1&91.01$\pm$0.3&90.43$\pm$0.6&90.34$\pm$0.8&90.35$\pm$0.8&90.66$\pm$0.5&91.59$\pm$0.2&\textbf{94.34$\pm$0.4}\\
    \cmidrule{2-14}
    &\multirow{3}{*}{AUC}
    &20&74.84$\pm$0.7&90.59$\pm$4.3&98.15$\pm$0.1&97.69$\pm$0.0&98.21$\pm$0.2&97.96$\pm$0.4&98.07$\pm$0.6&96.99$\pm$1.4&97.75$\pm$0.3&98.32$\pm$0.1&\textbf{99.16$\pm$0.1}\\
    &&40&78.54$\pm$0.6&91.42$\pm$4.0&97.85$\pm$0.1&97.08$\pm$0.0&97.93$\pm$0.1&97.70$\pm$0.3&97.48$\pm$0.6&97.12$\pm$0.4&97.23$\pm$0.2&98.06$\pm$0.1&\textbf{98.62$\pm$0.1}\\
    &&60&81.74$\pm$0.4&91.73$\pm$3.8&98.37$\pm$0.1&98.00$\pm$0.0&98.49$\pm$0.1&97.97$\pm$0.2&97.96$\pm$0.5&97.76$\pm$0.5&97.72$\pm$0.4&98.59$\pm$0.1&\textbf{99.33$\pm$0.1}\\
    
    \midrule
    \multirow{9}{*}{ACM}&
    \multirow{3}{*}{Mac-F1}
    &20&71.96$\pm$1.1&47.13$\pm$4.7&62.72$\pm$3.1&51.91$\pm$0.9&55.13$\pm$1.5&72.11$\pm$0.9&85.66$\pm$2.1&79.27$\pm$3.8&87.86$\pm$0.2&88.56$\pm$0.8&\textbf{89.90$\pm$0.9}\\
    &&40&73.76$\pm$0.8&55.96$\pm$6.8&61.61$\pm$3.2&62.41$\pm$0.6&61.21$\pm$0.8&72.02$\pm$0.4&87.47$\pm$1.1&80.23$\pm$3.3&86.23$\pm$0.8&87.61$\pm$0.5&\textbf{90.47$\pm$0.5}\\
    &&60&74.03$\pm$0.8&56.59$\pm$5.7&61.67$\pm$2.9&61.13$\pm$0.4&64.35$\pm$0.8&74.33$\pm$0.6&88.41$\pm$1.1&80.03$\pm$3.3&87.97$\pm$0.4&89.04$\pm$0.5&\textbf{90.15$\pm$0.4}\\
    \cmidrule{2-14}
    &\multirow{3}{*}{Mic-F1}
    &20&70.27$\pm$1.4&49.72$\pm$5.5&68.02$\pm$1.9&53.13$\pm$0.9&57.47$\pm$1.5&71.89$\pm$1.1&85.11$\pm$2.2&79.63$\pm$3.5&87.60$\pm$0.8&88.13$\pm$0.8&\textbf{89.63$\pm$1.0}\\
    &&40&73.14$\pm$1.0&60.98$\pm$3.5&66.38$\pm$1.9&64.43$\pm$0.6&62.62$\pm$0.9&74.46$\pm$0.8&87.21$\pm$1.2&80.41$\pm$3.0&86.02$\pm$0.9&87.45$\pm$0.5&\textbf{90.24$\pm$0.5}\\
    &&60&72.86$\pm$1.0&60.72$\pm$4.3&65.71$\pm$2.2&62.72$\pm$0.3&65.15$\pm$0.9&76.08$\pm$0.7&88.10$\pm$1.2&80.15$\pm$3.2&87.82$\pm$0.5&88.71$\pm$0.5&\textbf{89.89$\pm$0.4}\\
    \cmidrule{2-14}
    &\multirow{3}{*}{AUC}
    &20&86.31$\pm$0.8&65.88$\pm$3.7&79.50$\pm$2.4&71.66$\pm$0.7&75.44$\pm$1.3&84.36$\pm$1.0&93.47$\pm$1.5&91.47$\pm$2.3&96.72$\pm$0.3&96.49$\pm$0.3&\textbf{97.08$\pm$0.3}\\
    &&40&86.75$\pm$0.6&71.06$\pm$5.2&79.14$\pm$2.5&80.48$\pm$0.4&79.84$\pm$0.5&85.01$\pm$0.6&94.84$\pm$0.9&91.52$\pm$2.3&96.35$\pm$0.3&96.40$\pm$0.4&\textbf{97.49$\pm$0.3}\\
    &&60&88.11$\pm$0.6&70.45$\pm$6.2&77.90$\pm$2.8&79.33$\pm$0.4&81.64$\pm$0.7&87.64$\pm$0.7&94.68$\pm$1.4&91.41$\pm$1.9&96.79$\pm$0.2&96.55$\pm$0.3&\textbf{97.43$\pm$0.1}\\
    
    \midrule
    \multirow{9}{*}{AMiner}&
    \multirow{3}{*}{Mac-F1}
    &20&60.77$\pm$1.5&42.46$\pm$2.5&60.22$\pm$2.0&54.78$\pm$0.5&58.32$\pm$1.1&50.06$\pm$0.9&56.07$\pm$3.2&51.61$\pm$3.2&59.50$\pm$2.1&71.38$\pm$1.1&\textbf{72.91$\pm$1.0}\\
    &&40&67.64$\pm$1.1&45.77$\pm$1.5&65.66$\pm$1.5&64.77$\pm$0.5&64.50$\pm$0.7&58.97$\pm$0.9&63.85$\pm$1.5&54.72$\pm$2.6&61.92$\pm$2.1&73.75$\pm$0.5&\textbf{75.18$\pm$0.6}\\
    &&60&68.55$\pm$0.1&44.91$\pm$2.0&63.74$\pm$1.6&60.65$\pm$0.3&65.53$\pm$0.7&57.34$\pm$1.4&62.02$\pm$1.2&55.45$\pm$2.4&61.15$\pm$2.5&\textbf{75.80$\pm$1.8}&75.34$\pm$0.7\\
    \cmidrule{2-14}
    &\multirow{3}{*}{Mic-F1}
    &20&66.01$\pm$2.0&49.68$\pm$3.1&65.78$\pm$2.9&60.82$\pm$0.4&63.64$\pm$1.1&61.49$\pm$2.5&68.86$\pm$4.6&62.39$\pm$3.9&63.93$\pm$3.3&78.81$\pm$1.3&\textbf{80.20$\pm$0.8}\\
    &&40&73.05$\pm$1.3&52.10$\pm$2.2&71.34$\pm$1.8&69.66$\pm$0.6&71.57$\pm$0.7&68.47$\pm$2.2&76.89$\pm$1.6&63.87$\pm$2.9&63.60$\pm$2.5&80.53$\pm$0.7&\textbf{82.15$\pm$0.4}\\
    &&60&73.55$\pm$1.1&51.36$\pm$2.2&67.70$\pm$1.9&63.92$\pm$0.5&69.76$\pm$0.8&65.61$\pm$2.2&74.73$\pm$1.4&63.10$\pm$3.0&62.51$\pm$2.6&\textbf{82.46$\pm$1.4}&82.07$\pm$0.4\\
    \cmidrule{2-14}
    &\multirow{3}{*}{AUC}
    &20&86.18$\pm$0.9&70.86$\pm$2.5&85.39$\pm$1.0&81.22$\pm$0.3&83.35$\pm$0.5&77.96$\pm$1.4&78.92$\pm$2.3&75.89$\pm$2.2&85.34$\pm$0.9&90.82$\pm$0.6&\textbf{93.05$\pm$0.3}\\
    &&40&90.57$\pm$0.5&74.44$\pm$1.3&88.29$\pm$1.0&88.82$\pm$0.2&88.70$\pm$0.4&83.14$\pm$1.6&80.72$\pm$2.1&77.86$\pm$2.1&88.02$\pm$1.3&92.11$\pm$0.6&\textbf{94.81$\pm$0.2}\\
    &&60&90.71$\pm$0.5&74.16$\pm$1.3&86.92$\pm$0.8&85.57$\pm$0.2&87.74$\pm$0.5&84.77$\pm$0.9&80.39$\pm$1.5&77.21$\pm$1.4&86.20$\pm$1.7&92.40$\pm$0.7&\textbf{94.27$\pm$0.2}\\
    
    \midrule
    \multirow{9}{*}{Freebase}&
    \multirow{3}{*}{Mac-F1}
    &20&55.60$\pm$1.3&45.14$\pm$4.5&53.81$\pm$0.6&53.96$\pm$0.7&55.78$\pm$0.5&52.72$\pm$1.0&53.16$\pm$2.8&54.90$\pm$0.7&55.79$\pm$0.9&\textbf{59.23$\pm$0.7}&59.15$\pm$1.0\\
    &&40&57.58$\pm$1.2&44.88$\pm$4.1&52.44$\pm$2.3&57.80$\pm$1.1&59.28$\pm$0.6&48.57$\pm$0.5&59.63$\pm$2.3&53.40$\pm$1.4&49.88$\pm$1.9&61.19$\pm$0.6&\textbf{62.93$\pm$0.7}\\
    &&60&55.54$\pm$1.2&45.16$\pm$3.1&50.65$\pm$0.4&55.94$\pm$0.7&56.50$\pm$0.4&52.37$\pm$0.8&56.77$\pm$1.7&53.81$\pm$1.1&52.10$\pm$0.7&\textbf{60.13$\pm$1.3}&60.08$\pm$1.0\\
    \cmidrule{2-14}
    &\multirow{3}{*}{Mic-F1}
    &20&58.75$\pm$1.3&54.83$\pm$3.0&55.20$\pm$0.7&56.23$\pm$0.8&57.92$\pm$0.5&56.85$\pm$0.9&57.24$\pm$3.2&58.16$\pm$0.9&58.26$\pm$0.9&\textbf{61.72$\pm$0.6}&61.69$\pm$1.1\\
    &&40&60.59$\pm$1.2&57.08$\pm$3.2&56.05$\pm$2.0&61.01$\pm$1.3&62.71$\pm$0.7&53.96$\pm$1.1&63.74$\pm$2.7&57.82$\pm$0.8&54.28$\pm$1.6&64.03$\pm$0.7&\textbf{65.99$\pm$1.0}\\
    &&60&58.44$\pm$1.2&55.92$\pm$3.2&53.85$\pm$0.4&58.74$\pm$0.8&58.57$\pm$0.5&56.84$\pm$0.7&61.06$\pm$2.0&57.96$\pm$0.7&56.69$\pm$1.2&63.61$\pm$1.6&\textbf{63.82$\pm$1.5}\\
    \cmidrule{2-14}
    &\multirow{3}{*}{AUC}
    &20&73.20$\pm$1.1&67.63$\pm$5.0&73.03$\pm$0.7&71.78$\pm$0.7&73.89$\pm$0.4&70.84$\pm$0.7&73.26$\pm$2.1&72.80$\pm$0.6&73.19$\pm$1.2&76.22$\pm$0.8&\textbf{76.96$\pm$0.8}\\
    &&40&75.25$\pm$1.0&66.42$\pm$4.7&74.05$\pm$0.9&75.51$\pm$0.8&76.08$\pm$0.4&69.48$\pm$0.2&77.74$\pm$1.2&72.97$\pm$1.1&70.77$\pm$1.6&78.44$\pm$0.5&\textbf{79.25$\pm$0.6}\\
    &&60&74.20$\pm$1.4&66.78$\pm$3.5&71.75$\pm$0.4&74.78$\pm$0.4&74.89$\pm$0.4&71.01$\pm$0.5&75.69$\pm$1.5&73.32$\pm$0.9&73.17$\pm$1.4&78.04$\pm$0.4&\textbf{78.32$\pm$0.7}\\
  \bottomrule
\end{tabular}}
\end{table*}

\subsubsection{Implementation Details.} 
In the settings of our method, we did not use any meta-paths or meta-graphs.
For $\text{COM}$ in Eq. (\ref{message}), we used summation for DBLP and ACM, and concatenation for AMiner and Freebase.
For the direct heterogeneous neighbor aggregation in Eq. (\ref{message}), we assigned GraphSAGE \cite{hamilton2017inductive} modules as many numbers as the edge types in the dataset to compute $h_{i,1}^{(l)}, \ldots, h_{i,R}^{(l)}$.
When computing the vector representation of meta-nodes $h^{(l)}_{\phi(v_i)}$ in Eq. (\ref{message}), every node messages in each node type was aggregated.
However, aggregating all the information about each node type into one fixed-size vector can lead to over-squashing issues and lose its intended meaning \cite{alon2020bottleneck, topping2021understanding}.
To solve this issue, for each training epoch, we randomly connect only $r \%$ nodes in each node type to each meta-node, where $r \in \{40, 50, 60, 70, 80, 90\}$.
The choice of hyper-parameter and further implementation details are described in the supplementary material for reproducibility.
\begin{table}[t]
\centering
\caption{Summary of node clustering results ($\%$).}
  \label{tab:clustering}
  \resizebox{0.46\textwidth}{!}{\begin{tabular}{lcccccccc}
    \toprule
      & \multicolumn{2}{c}{DBLP} & \multicolumn{2}{c}{ACM} & \multicolumn{2}{c}{AMiner} & \multicolumn{2}{c}{Freebase} \\
\cmidrule{2-9}
& NMI & ARI & NMI & ARI & NMI & ARI & NMI & ARI \\
\midrule
    n2vec & 21.48&14.70&41.71&34.77&32.04&14.36&16.43&17.27\\
    SAGE & 51.50&36.40&29.20&27.72&15.74&10.10&9.05&10.49\\
    GAE & 72.59&77.31&27.42&24.49&28.58&20.90&19.03&14.10\\
    mp2vec & 73.55&77.70&48.43&34.65&30.80&25.26&16.47&17.32\\
    HERec & 70.21&73.99&47.54&35.67&27.82&20.16&19.76&19.36\\
    HetGNN & 69.79&75.34&41.53&34.81&21.46&26.60&12.25&15.01\\
    DGI &  59.23&61.85&51.73&41.16&22.06&15.93&18.34&11.29\\
    DMGI & 70.06&75.46&51.66&46.64&19.24&20.09&16.98&16.91\\
    HeCo & 74.51&80.17&56.87&56.94&32.26&28.64&\textbf{20.38}&\textbf{20.98}\\
    \midrule
    MN (ours) & \textbf{78.39}&\textbf{83.02}&\textbf{63.56}&\textbf{67.35}&\textbf{37.69}&\textbf{29.05}&17.13 &18.39 \\
  \bottomrule
\end{tabular}}
\end{table}

\subsection{Node Classification}
We conducted node classification to see how useful the learned representation from the meta-node message passing encoder of contrastive learning is.
For each dataset, we selected $20, 40, 60$ nodes per class for training set, $1,000$ nodes for validation set, and $1,000$ nodes for test set.
We trained and tested a single layer of logistic regression classifier, and used Macro-F1, Micro-F1, and AUC for evaluation metrics.
The average value and standard deviation after executing each model $10$ times are reported in Table \ref{tab:classification}.
The results demonstrate that our method (MN) can achieve outstanding results compared to the existing homogeneous models and heterogeneous models even without any predefined composition of heterogeneous nodes such as meta-paths.
Especially, in most cases, the proposed method shows outperforming results compared to state-of-the-art heterogeneous models (mp2vec, DMGI, HeCo, etc.) that rely on the pre-configured meta-paths.
Also, it can be seen that our method shows outstanding performances compared to contrastive learning models including DGI, DMGI, and HeCo.
We have also observed that, for AMiner and Freebase datasets where the node feature does not have proper information about the semantics of the node, homogeneous models can achieve similar performances to heterogeneous models.
Specifically, n2vec and GAE show classification performances close to those of several heterogeneous models such as mp2vec, HERec, HetGNN.
We conjecture that the node feature with rich semantics plays an important role in distinguishing different types of nodes of heterogeneous graphs.

\subsection{Node Clustering}
We conducted node clustering by applying k-means clustering algorithm to the learned representation of each model.
The clustering performance is measured by Normalized Mutual Information (NMI) and Adjusted Rand Index (ARI).
Table \ref{tab:clustering} reports the average value after executing each model $10$ times to consider random initialization of k-means clustering algorithm.
For most cases, the proposed method shows outstanding performance compared to the state-of-the-art.
The results of DMGI, HeCo, and our method demonstrate that the contrastive learning framework is effective to learn representations of heterogeneous graphs in unsupervised environments. 
We observed that every model shows poor performance on Freebase compared to other datasets.
Similar to \cite{fu2020magnn}'s analysis on IMDB movie dataset, we guess the cause of this result comes from the noisy labels of the movie genres.
Every movie can have multiple genres, but for the classification task, only one genre was selected as a label among them.
As evidence for this conjecture, we found that another paper \cite{li2016transductive} used different movie genre labels, \textit{Action, Adventure}, and \textit{Crime} for the Freebase dataset, while, in our experiments, we used \textit{Action, Comedy} and \textit{Drama} labels. 

\begin{figure*}[t!]
     \centering
     \begin{subfigure}[b]{0.3\textwidth}
         \centering
         \includegraphics[width=\textwidth]{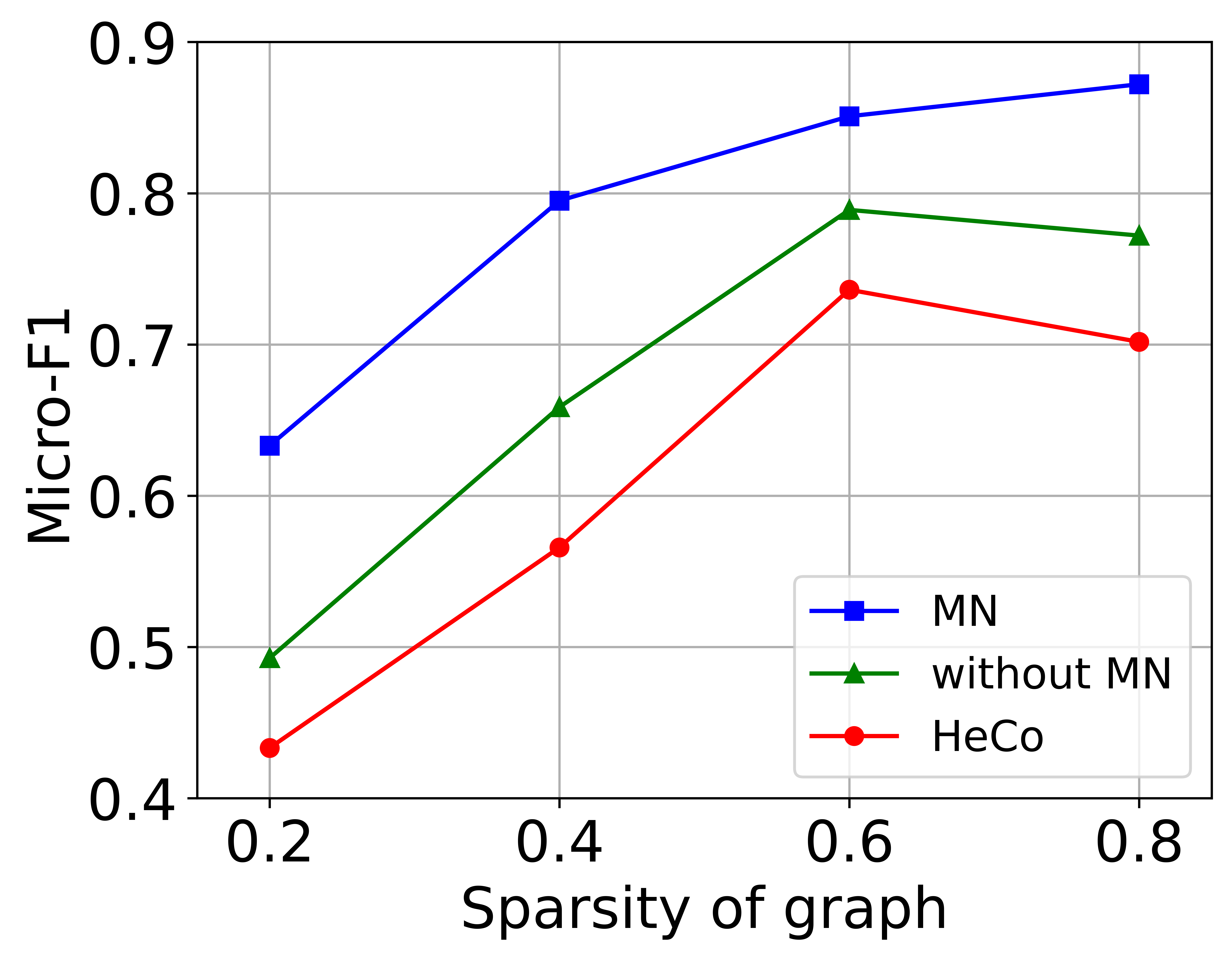}
         \caption{ACM}
     \end{subfigure}
     \hfill
     \begin{subfigure}[b]{0.3\textwidth}
         \centering
         \includegraphics[width=\textwidth]{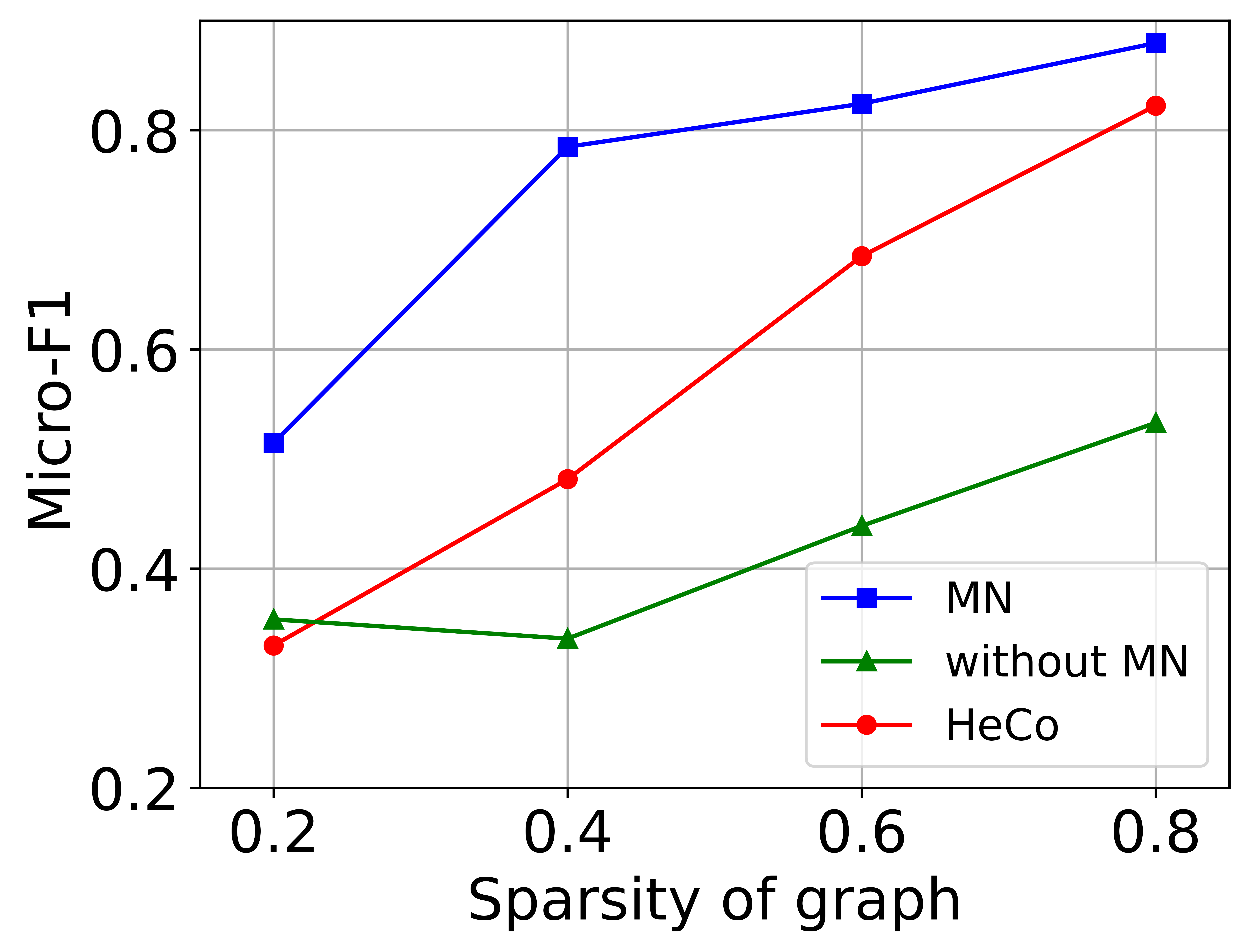}
         \caption{DBLP}
     \end{subfigure}
     \hfill
     \begin{subfigure}[b]{0.3\textwidth}
         \centering
         \includegraphics[width=\textwidth]{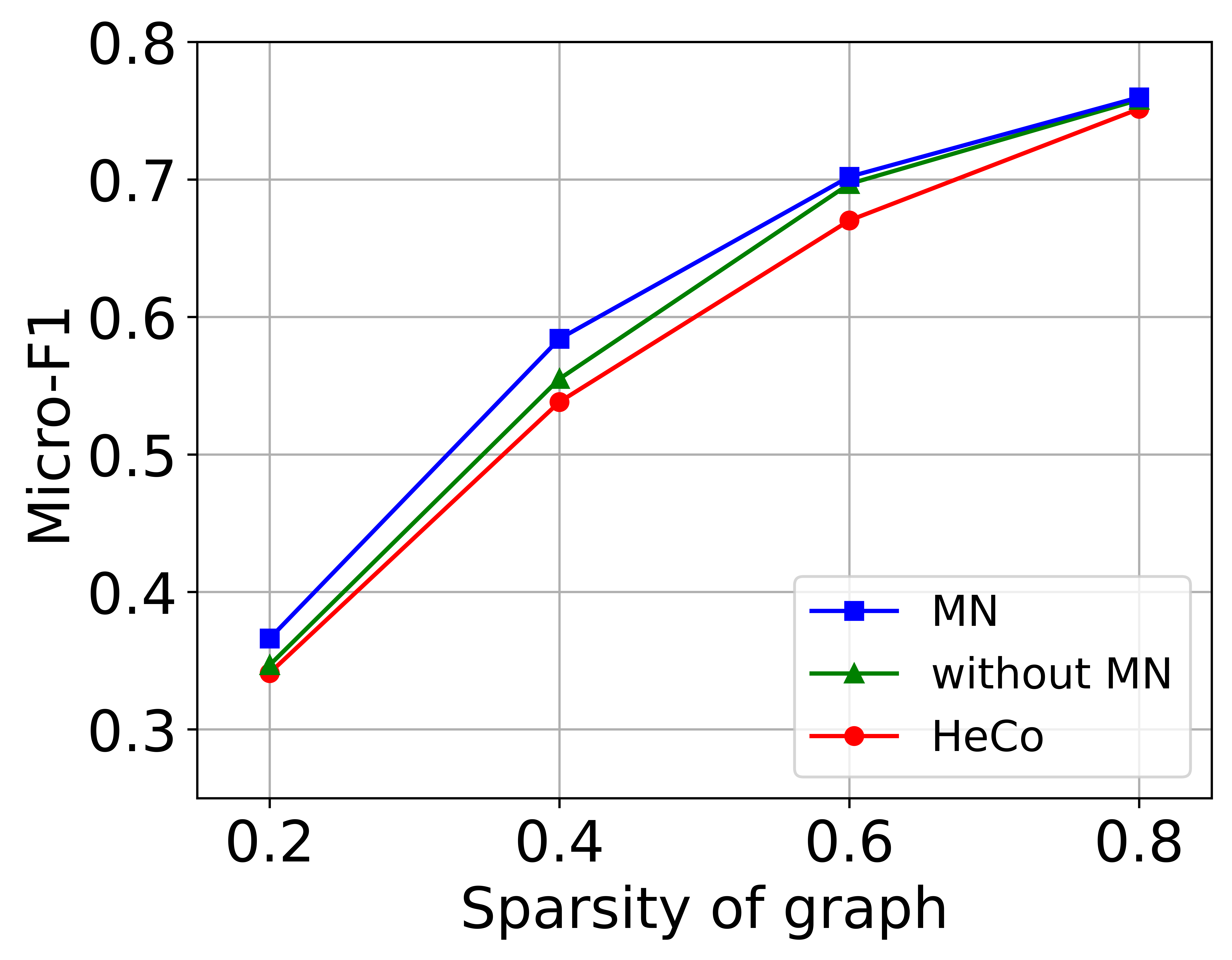}
         \caption{AMiner}
     \end{subfigure}
        \caption{
        The classification results of sparse graphs when the training set is composed of $40$ nodes per each class.
        MN (blue line), without MN (green line), and HeCo (red line) denote the proposed message passing scheme, the message passing that only considers the message of each node and the aggregated messages from heterogeneous neighbors, and the message passing scheme that relies on the meta-paths.
        For instance, $0.2$ of the sparsity of graph indicates there exists only $20 \%$ of original edges.
        }
        \label{fig:sparse}
\end{figure*}

\subsection{Effectiveness of Meta-nodes} 
If there are few inter-type relations, then existing heterogeneous models cannot learn the intra-type relations well due to the limited number of meta-paths that are composed of inter-type relations.
To show the effectiveness of meta-nodes compared to meta-paths, we make the graphs of ACM, DBLP, and AMiner sparse by randomly removing a fraction of the edges.
Then, we measured the node classification performances of three models: i) the proposed model (MN), ii) aggregating only messages of each node and direct heterogeneous neighbors without using meta-node representation (without MN), and iii) HeCo which relies on meta-paths.
The results are presented in Figure \ref{fig:sparse}.
In the results of ACM and DBLP, by comparing MN and without MN, it can be noticed that the proposed meta-node message passing scheme enables learning enriched relational knowledge by leveraging both inter- and intra-type relations effectively.
Also, due to the decreased number of meta-paths by sparsifying graphs, the performance of HeCo deteriorated severely.
In the case of ACM, the performance of without MN is better than that of HeCo.
We conjecture that both view masking mechanism and positive sample mining that utilize meta-paths in HeCo are significantly affected by the reduced number of meta-paths in some cases.
On the other hand, every method shows similar performances on AMiner.
This is because, as shown in Table \ref{tab:data}, all the given edges are connected to the target node type and are abundant compared to the number of target nodes.
Therefore, if the inter-type relations connecting the target type nodes are abundant, target nodes can aggregate enough information, or one can create a sufficient number of meta-path.

\subsection{Qualitative Analysis}
\begin{figure}[ht!]
\captionsetup[subfigure]{justification=centering}
\centering
\rotatebox[origin=c]{90}{\makebox[0.5in]{DBLP}}%
  \quad
\begin{subfigure}{0.28\columnwidth}
\centering
\includegraphics[clip, trim={1.2cm 1.0cm 0.5cm 0.6cm}, width=\columnwidth]{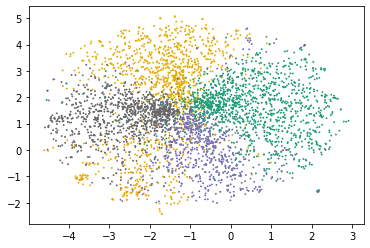}
\caption{n2vec: 0.1944}
\end{subfigure}\hfill
\begin{subfigure}{0.28\columnwidth}
\centering
\includegraphics[clip, trim={1.4cm 1.0cm 0.5cm 0.6cm}, width=\columnwidth]{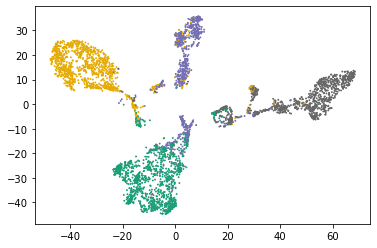}
\caption{HeCo: 0.5205}
\end{subfigure}\hfill
\begin{subfigure}{0.28\columnwidth}
\centering
\includegraphics[clip, trim={1.4cm 1.0cm 0.5cm 0.6cm}, width=\columnwidth]{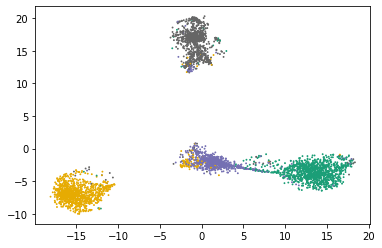}
\caption{MN: 0.6351}
\end{subfigure}\par

\rotatebox[origin=c]{90}{\makebox[0.5in]{ACM}}%
  \quad
\begin{subfigure}{0.28\columnwidth}
\centering
\includegraphics[clip, trim={1.4cm 1.0cm 0.5cm 0.6cm}, width=\columnwidth]{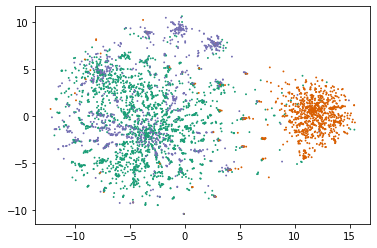}
\caption{n2vec: 0.2035}
\end{subfigure}\hfill
\begin{subfigure}{0.28\columnwidth}
\centering
\includegraphics[clip, trim={1.4cm 1.0cm 0.5cm 0.6cm}, width=\columnwidth]{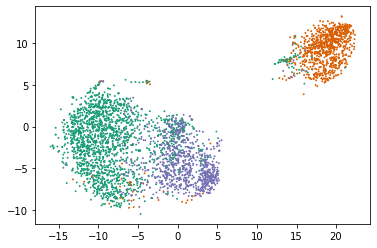}
\caption{HeCo: 0.4436}
\end{subfigure}\hfill
\begin{subfigure}{0.28\columnwidth}
\centering
\includegraphics[clip, trim={1.3cm 1.0cm 0.5cm 0.6cm}, width=\columnwidth]{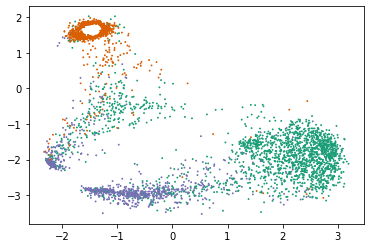}
\caption{MN: 0.4743}
\end{subfigure}\par

\rotatebox[origin=c]{90}{\makebox[0.5in]{AMiner}}%
  \quad
\begin{subfigure}{0.28\columnwidth}
\centering
\includegraphics[clip, trim={1.4cm 1.0cm 0.6cm 0.6cm}, width=\columnwidth]{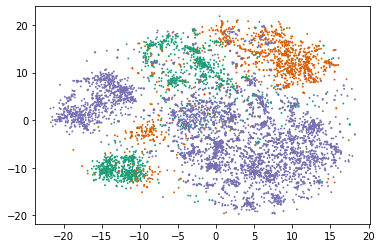}
\caption{n2vec: 0.0037}
\end{subfigure}\hfill
\begin{subfigure}{0.28\columnwidth}
\centering
\includegraphics[clip,trim={1.4cm 1.0cm 0.5cm 0.6cm},  width=\columnwidth]{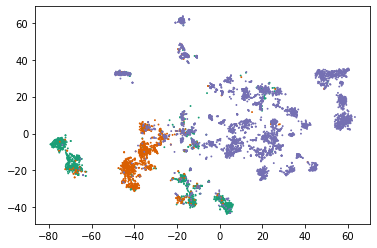}
\caption{HeCo: 0.1681}
\end{subfigure}\hfill
\begin{subfigure}{0.28\columnwidth}
\centering
\includegraphics[clip, trim={1.4cm 1.0cm 0.5cm 0.6cm}, width=\columnwidth]{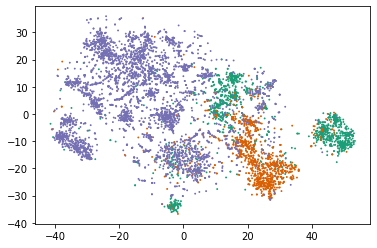}
\caption{MN: 0.1692}
\end{subfigure}\hfill
\caption{
The $2$D projections of learned representations of n2vec, HeCo, and our method (MN) for DBLP, ACM, and AMiner are illustrated.
Silhouette scores of each projected representation are provided below each subfigure.
}
\label{fig:vis}
\end{figure}

For qualitative analysis of the proposed method, we projected the learned representation of the target node type of each dataset.
By using t-SNE \cite{van2008visualizing}, we obtained $2$D projections of learned representations.
The projection results are shown in Figure \ref{fig:vis}.
Nodes of the same color share the same label.
To measure the quality of projected representations of each model, we measured a silhouette score \cite{rousseeuw1987silhouettes}.
Our method can distinguish node classes better than comparison methods.
The projected representations of n2vec overlap a lot among other classes due to the limited learning ability of homogeneous models.
The projected representations of HeCo show discriminability among classes.
However, these results can only be achieved with well-configured meta-paths in advance. 

\section{Conclusions}
In order to comprehensively learn complex relations in heterogeneous graphs, we proposed a concise and powerful concept of meta-node from the understanding of the unique structural characteristics of heterogeneous graphs.
By introducing meta-nodes, each node can take into account information both of heterogeneous neighbors and its corresponding node type by explicit modeling of intra-type relations.
Then, we proposed a meta-node message passing layer (MN-MPL) and applied MN-MPL to the contrastive learning model.
The proposed method was validated qualitatively and quantitatively on real-world heterogeneous graphs.
Unlike meta-paths and meta-graphs, the proposed meta-node does not require expert knowledge to construct, does not need to be designed before learning, and can learn well enough even if there are only a small number of edges.
Our results shed light on the possibility of accurate heterogeneous graph learning even without meta-paths or meta-graphs which are frequently used in this field.
In the future, we will take one step further about message passing using meta-nodes.
Through meta-node message passing, messages can be received from nodes of the same type, but not all of the same type have the same importance.
Thus, computation of importance among nodes of the same type by self-attention mechanism or shortest path distance is one of the promising directions.


\bibliography{aaai23}

\end{document}